# Image classification based on support vector machine and the fusion of complementary features

**Huilin Gao, [a,b] Wenjie Chen, [a,b,*] Lihua Dou, [a,b]**
[a]Beijing Institute of Technology, School of Automation, No. 5 South Zhongguancun Street, Beijing, China, 100081
[b]State Key Laboratory of Intelligent Control and Decision of Complex System, No. 5 South Zhongguancun Street, Beijing, China, 100081

**Abstract**. Image Classification based on BOW (Bag-of-words) has broad application prospect in pattern recognition field but the shortcomings are existed because of single feature and low classification accuracy. To this end we combine three ingredients: (i) Three features with functions of mutual complementation are adopted to describe the images, including PHOW (Pyramid Histogram of Words), PHOC (Pyramid Histogram of Color) and PHOG (Pyramid Histogram of Orientated Gradients). (ii) The improvement of traditional BOW model is presented by using dense sample and an improved K-means clustering method for constructing the visual dictionary. (iii) An adaptive feature-weight adjusted image categorization algorithm based on the SVM and the fusion of multiple features is adopted. Experiments carried out on Caltech 101 database confirm the validity of the proposed approach. From the experimental results can be seen that the classification accuracy rate of the proposed method is improved by 7%-17% higher than that of the traditional BOW methods. This algorithm makes full use of global, local and spatial information and has significant improvements to the classification accuracy.

**Keywords**: image classification, complementary features, Bag-of-words (BOW), feature fusion, support vector machine (SVM).

\* Wenjie Chen, E-mail: chen.wenjie@163.com

## 1 Introduction

Since the age of internet was coming, it is becoming more and more convenient and quick to acquire and transmit the digital image. With the popularity of digital cameras and camera-phones, the rapid promotion of 3G wireless network bandwidth and the continuous development of online business such as e-commerce, digital images are all around us all the time. Image has been an important information carrier of today's internet field. Efficient image classification techniques are required urgently with the rapid increase and potential huge use of image resources. In the field of computer vision, image classification has become heatedly discussed topics in recent years.[1] How to classify the images is an urgent practical problem to be solved in the face of huge image information.

Image classification differs from traditional problem of image retrieval: images are classified under particular fields and classification systems which are generally predefined, such as classification problems of sport images within the Olympic Movement. In particular, Image classification is to give the category information for an image.

The content-based image classification is to do some semantic type classification to the images according to their visual features, which is a hot research topic at present. Firstly, visual features are extracted to represent and describe the image accurately employing image processing and analysis technique, that is, the extraction of object and image representation. And then utilizing machine learning algorithms to build classification models which are used to classify the testing image. Youna adopted color and texture feature to classify the moving images with Bayes classifier.[2] Chang and Goh use SVM and the global feature of the image to build classification model so as to realize the image classification.[3] Image representation based on bag-of-words model is the most popular method at present. SIFT local features were coded into visual words to construct the BOW model which is combined with SVM to categorize the images by Csurka.[4] The complex, cluttered local features are expressed as the regular and orderly visual words histogram vectors by BOW model. The BOW algorithm which is simply to implement has a good effect in image classification. However, global information of the image is easily overlooked, which leads to a negative impact on the classification accuracy. And in the practical applications of image classification, in view of the characteristics of image data itself, it is difficult for single feature to describe an image fully. The extraction and representation of image's visual information are important bottleneck restricting the performance of image classification and influencing the accuracy of the classification. So it is crucial to obtain complete and stable features accurately.

The paper puts emphases on several aspects of improvement as follows. (i) First the image representation: in the traditional bag-of-words method, the SIFT algorithm[5,6] is used to extract the feature points of images and the obtained SIFT features are used to generate a fixed number of the typical local features as visual words, which did not take full advantage of the spatial distribution of image information and the global information. It is exactly based on this consideration that the paper proposes to use patch-based global and local features with spatial pyramid matching to encode its spatial information for image feature extraction and description. Three features with functions of mutual complementation are adopted to extract and describe the features of images, including PHOW (Pyramid Histogram of Words), PHOC (Pyramid Histogram of Color) and PHOG (Pyramid Histogram of Orientated Gradients). The method considers both the global features(color and shape) and the local distribution information of the image, furthermore it is much more complete and flexible to describe the feature information of the image through multi-feature fusion and the pyramid structure composed by image spatial multi-resolution decomposition, which improves the accuracy of image classification. (ii) About the PHOW algorithm, the improvement of traditional BOW model is presented by using dense sample based on patches instead of the original sparse sample and an improved K-means clustering method for constructing the visual dictionary to overcome the influence on the clustering performance caused by the selection of initial value. (iii) On the design of classifier: for the extracted features (PHOW, PHOC and PHOG), due to the importance of each feature in the image is not exactly the same, an adaptive feature-weight adjusted image classification algorithm based on the SVM and the fusion of multiple features is adopted to avoid the impact brought by various dimension of each feature. This algorithm can automatically learn the weight coefficient of various features to effectively solve the defects that various features are simply assembled into one feature vector, which improves the performance of image classification.

The remainder of this paper is organized as follows: We introduce the feature descriptors and how the images are represented in detail in Sec. 2. Then the algorithm of feature fusion based on SVM classifier and multi-feature fusion are mainly elaborated in Sec. 3. As a result Sec. 4 shows a description of data sets and the experimental evaluation of the performance on Caltech-101, as well as a comparison with the state of the art. The paper concludes with a discussion and conclusions in Sec. 5.

## 2  Image Representation

Usually, in the image classification system, not only does single feature fail to represent the content of an image accurately, it also far from meet users' requirement for the classification effect of a variety of images. Therefore, the researchers generally extract features in various types from the image to form the feature vector for the content of an image. The different features of the image have complementary information, so we combine the global feature with local feature extraction in order to achieve higher classification accuracy.

For feature extraction this paper adopts three methods including PHOW[7], PHOC[8] and PHOG[9]. All three are formed as the through spatial multi-resolution decomposition of the image. Then the images are represented in a multi-resolution pyramid structure.[10] Color features are extracted and described by PHOC method, which considers the global information of the image. HOG features are extracted and described by PHOG method, which considers image feature information about the shape. SIFT features are extracted and described by PHOW method, which considers the local information feature of the image and solves the vulnerability that HOG feature and color feature are sensitive to rotation of the object.

*2.1 Pyramid Histogram of Words（PHOW） with Improved BOW*

The main idea of Bag-of-words method is as follows: Firstly, extract the local features from all images; Then, cluster the feature with K-means algorithm to form several clustering centers which are the typical local features known as the "visual keywords" (code words) to compose visual vocabulary (code); Finally, the local features extracted from each image can then be mapped to to the key words in the vocabulary and the frequency for the visual keywords (the number of times each keyword appears) in each image are counted to get the histogram as the image BOW representation. [11-13]

The method of local features description based on BOW obtained excellent performance in image classification.[11] However, there are several defectiveness and problems with the traditional BOW model: (i) The method of sparse sampling is not favorable to improve the performance of object classification; (ii) In the process of generating visual keywords, the traditional K-means algorithm has its limitations that initialization of clustering had a great influence on the clustering results. If the location of the $V$ initial centers was too centralized or unevenly distributed throughout the whole data space, it would have caused the computational burden because of the increase in the number of iterations required for convergence and computational complexity increases or trapped into local optimum so that we can't get accurate clustering results; (iii) The spatial location information of the image was ignored. Considering the above reasons, we improved the traditional BOW method from three aspects as follows.

(1) The dense sampling[14] based on block is adopted to obtain a kind of densely extracted SIFT descriptor for feature extraction. First each image is partitioned into uniform grid patches by uniform sampling. In this way, each image is represented by a number of patches and each patch is represented by a SIFT feature vector.

(2) Results obtained from lots of experiments on K-means clustering algorithm demonstrate that, the selection of the initial point has effect on the final convergence results to some extent. So if we can find a suitable initial point selection algorithm, it would not only reduce the running time and iterations of clustering algorithm but also improve the classification accuracy. The traditional K-means clustering methods used random approach to obtain the distribution of initial centers for iteration. Here we use the improved version K-means++[15] which is a method of choosing probability. The specific steps of the algorithm are as follows:

a) Select an initial center $c_1$ randomly from the sample set $X$ as the beginning of a cluster center;

b) Select a new center $c_i = x' \in X$ from the sample set $X$ and make it meet the probability distribution $\dfrac{D(x')^2}{\sum_{x \in X} D(x)^2}$ in which $D(x)$ indicates the shortest distance between the data point x and the selected clustering center that is nearest to it;

c) Repeat step b) until $V$ eligible centers are found and the resulting $V$ initial centers overcome the influence on the clustering performance caused by random selection.

d) Perform traditional K-means algorithm on the basis of the above.

(3) Traditional BOW models ignored the spatial characteristics of the image, thus an improved model named PHOW (Pyramid Histogram of Words) as shown in Ref. 7. The image is partitioned level by level and each level of image is composed of several blocks. Then a series of visual words histograms are formed for the representation of image from low resolution to high resolution in the feature space. The establishment process of the PHOW model are as follows:

a) The hierarchical image pyramids have been built for grid partitioning in the first place. Partition the source images hierarchically into a quad-tree of multilevel sub-blocks.

Suppose the highest level recorded as L is 3. The 1st level (recorded as l = 0) is the whole image, the 2nd level(l = 1) is obtained by partitioning the 1st level into quarters, the 3rd level (l = 2) is obtained by partitioning each sub-block of the 2nd level into quarters, and so on. The maximum level is set as $L$. The image would be partitioned into $4^l$ sub-areas at the $l$th level of the pyramid. The hierarchical partition result of spatial pyramid for the image is shown in Fig. 1(a).

b) Calculate the feature histograms of the sub-areas at all different levels of the image. After the dense sampling for all the sub-areas at all different levels of the image, we extract the SIFT features from its surrounding area for each grid sampling point to describe it. And then K-means++ is used to cluster the feature points to obtain the final cluster centers as the visual words. Each sub-area will generate a visual vocabulary composed of $V$ visual words. By the statistics of the histograms of all sub-areas at each level, we can obtain the feature histogram of the image at each level.

c) These different feature histograms of the image at different levels are assigned with corresponding weightings (the histogram at $l$th level of mergers of gradient direction normalized weight) and are concatenated into a one-dimensional vector as the PHOW model of the image.

If the number of the bin about the histogram of each sub-area is $V$ and the highest level is $L$, then the dimension of the feature vector of the image at the $l$th level can be computed by $V \sum_{l=0}^{L} 4^l$.[16] The result of the PHOW is shown as Fig. 1(b).

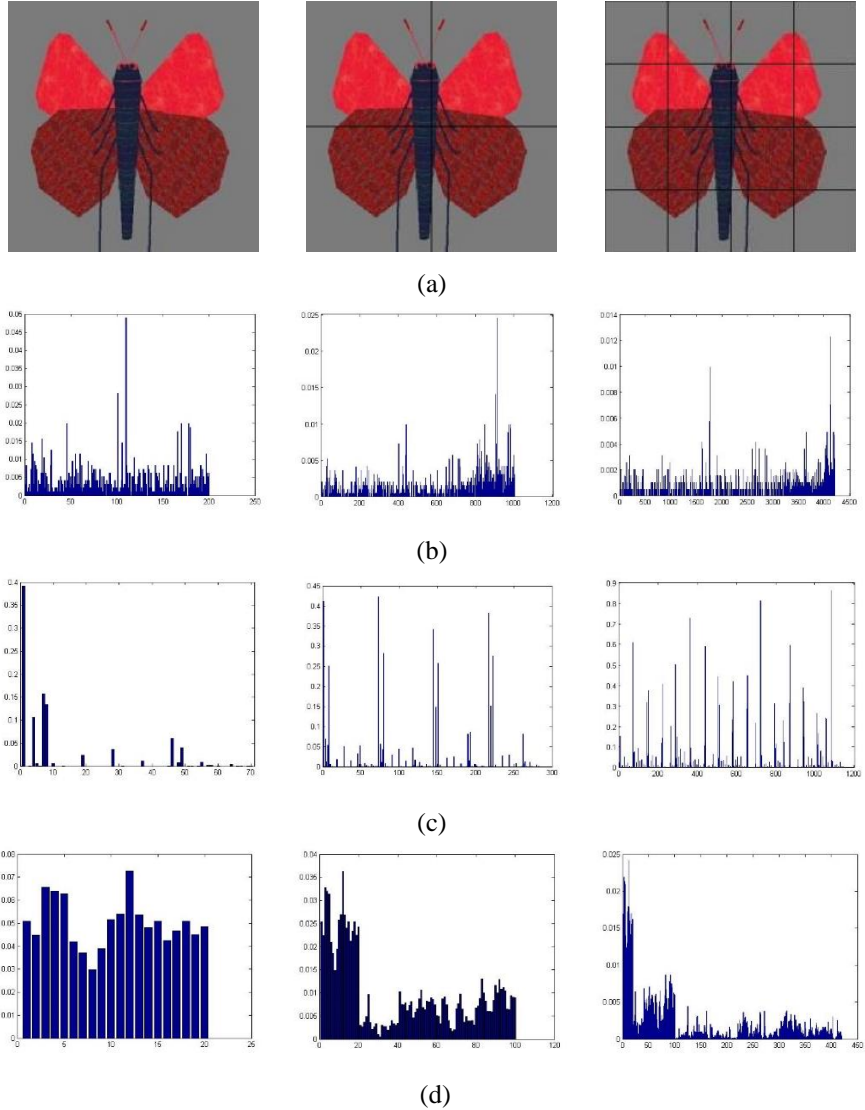

**Fig. 1** PHOW、PHOC and PHOG representation: (a) Grids for levels l = 0 to l = 2 of the Pyramid space hierarchy, (b) PHOW representation corresponding to each level, (c) PHOC representation corresponding to each level and (d) PHOG representation corresponding to each level.

## 2.2 Pyramid Histogram of Color（PHOC）

Color histogram is an important and widely used image feature with simple calculation and good robustness which can express the information of global image effectively and is not sensitive to location, direction and geometric size of the object.[17] In all of the image color space, HSV color space is a uniform color space which can well reflect people's perception ability of color and its components indirectly affect the visual perception. So we adopt HSV histogram. The color signals are represented in HSV as three attributes: Hue, Saturation and Value, in which "Hue" indicates different colors , "Saturation" indicates the shade of color and "Value" indicates the brightness of color mainly affected by the intensity of the light source. However, due to an image is multi-colored, so it is intolerable with both the storage space and computation time for directly performing computations in three-dimensional color space. Therefore, HSV space must be properly quantified before the calculation of the histogram.That is, the whole color space is divided into several sub-spaces, each subspace corresponds to one bin of the histogram, then color histogram is obtained by counting the number of pixels which lie within the subspace corresponding to each bin of the histogram.

Color histogram shows the number of features at all levels about the color of image which can reflect the part information of the image while it ignores the spatial location information of the image so it is possible for different images to have the same color histogram. Based on this we adopted PHOC feature for experiment. The main idea is to use spatial quad-tree partitions to form the multi-resolution representations of the image and describe the image by the combination of the multi-level color histograms from low-resolution representation to high-resolution representation of the image. In the paper, the concrete steps of the PHOC feature extraction are as follows:

(1) The hierarchical image pyramids have been built for grid partitioning in the first place. This is the same as Step a) in Part (3) of sec. 2.1.

(2) Extract the HSV color histogram in each sub-block of images in all partition levels and assign different weights for the histogram of each level.

(3) These color histograms at all levels with different weights are combined into one vector as the PHOC feature representation of image for training and classification. The formula of PHOC[8] is indicated as follows:

$$H(n_i) = \frac{1}{2^L}\frac{n_i}{N}, l = 0 \tag{1}$$

$$H(n_i) = \frac{1}{2^{L-l+1}}\frac{n_i}{N}, l > 0 \tag{2}$$

Where $n_i$ denotes the number of pixels in the ith bin of the color space, $N$ denotes the number of pixels in the whole image, $L$ is the highest level, $l$ is the current level, $\frac{1}{2^L}$ is the weight of the level $l = 0$, $\frac{1}{2^{L-l+1}}$ is the weight of the level $l > 0$. The result of the PHOC is shown as Fig. 1(c).

*2.3 Pyramid Histogram of Oriented Gradient（PHOG）*

HOG (histogram of oriented gradient descriptor) is a feature descriptor for target detection applied in the field of computer vision and image processing. Navneet Dalal and Bill Triggs proposed HOG for pedestrian detection in static images or videos.[18] The technique counts the number of times the oriented gradient appeared in the local image. It is an effective method to describe shape information of the image. Through extracting the information about distribution of edge or gradient for the local area of the image which can be a good characterization of the structure about the edge or gradient of the object in local area, thus achieving to represent the shape of the object. Firstly, the whole image is segmented into dense grid of uniformly spaced cells and the gradient direction is divided into K bins. Then the gradient of all pixels in each cell are computed to generate a histogram of oriented gradient (composed of the number of different gradients) which would be expressed as a $K$-dimension feature vector to form the descriptor of each cell. At last, the descriptors of all cells are interconnected to represent the feature vector of the sub-images. The HOG descriptor has a good robustness against illumination and geometrical changes.

HOG has actually been considering the spatial distribution information of image, but doesn't take into account the impact on classification performance caused by spatial scale division in different ways to the image. Therefore, Bosch proposed pyramid histogram of orientated gradients (PHOG)[9] as a shape feature[16] which can better describe both local shape and spatial distribution of the objects. Firstly, the image is hierarchically partitioned into a quad-tree of multilevel sub-blocks to generate the image pyramid. And then extract HOG feature from each cell in every level of the image pyramid. Finally, combine multistage HOG of different levels from low resolution to high resolution to form the PHOG feature of the image. With the increasing of partition level, it is more and more refined and localizing for PHOG to describe the edge and shape of the image. The concrete steps of the PHOG feature extraction are as follows:

(1) The hierarchical image pyramids have been built for grid partitioning in the first place.. Partition the source images hierarchically into a quad-tree of multilevel sub-blocks. This is the same as Step a) in Part (3) of sec. 2.1.

(2) Extract the contour edge of the images in all partition levels by edge detection algorithm for shape description.

(3) Calculate gradient direction and amplitude of each edge pixel and the scope of gradient direction $\theta(i,j)$ is $[0, 360°]$ or $[0, 180°]$ which should be divided into $K$ intervals. Figures out the number of pixels whose gradient direction $\theta(i,j)$ take values in each interval and accumulate the gradient amplitudes of all pixels in each interval as the weight of the corresponding interval represented in the gradient direction histogram as shown in Eq. (3).

$$M(i,j) = \sqrt{I_i^2 + I_j^2}, \tag{3}$$

$$\theta(i,j) = \tan^{-1}\frac{I_j}{I_i} \in [0, 360°]. \tag{4}$$

(4) HOG of the images in all partition levels are normalized that the normalized weight of histograms of oriented gradient in the $l$th level is $\frac{1}{2^{L-l+1}}$. Sequentially concatenate these histograms to attain the final PHOG shape feature description. The result of the PHOG is shown as Fig. 1(d).

## 3　The design of classifier based on multi-feature fusion

The research of methods of feature fusion include the low-level feature fusion, the middle-level feature fusion and the high-level feature fusion. The problems associated with widely varying dimensions among different feature vectors and dimensions overflow will reduce the accuracy rates with the low-level and middle-level feature fusion method. The high-level feature fusion is a process of making decisions together according to the decision information for the corresponding categorized result with each feature. It is assumed that there are $N$ categories of image sets in the image database which is divided into two parts: training set $I$ and testing set $\hat{I}$ And each category of the training set is represented as $\{I_i^j\}_{i=1}^{N_j}, j = 1,\cdots,N$. Then $M$ different features are extracted from each image expressed as $F_m(I_i^j), m = 1, 2,\cdots, M$, that is, there are $M$ training sets per category. SVM classifier is designed to classify the feature vectors in each training set, and the best parameter that correspond to each set is obtained by cross-validation method. For the testing set $\hat{I}$, according to the classifier parameters based on each feature obtained in the training set, the probability that the image belongs to each category is expressed as $D_j(m), m=1,2,\cdots M$. Since each image has $M$ different features, the results of the classification is expressed as the set $\{D_j(m)\}_{m=1}^{M}$. The integration formula[19] is as follow:

$$j^* = \arg\max_{j=1}^{N}\left\{\sum_{m=1}^{M}a_m D_j(m)\right\}, \sum_{m=1}^{M}a_m = 1 \quad (5)$$

Where the weighing coefficients $a_m$ is the parameter obtained by the cross-validation method during the classification of the training set.

There is no such issue using the high-level feature fusion method which is just the fusion for final decision information rather than the features themselves.[19] Based on this, the high-level feature fusion has been used in the article, as shown in Fig. 2.

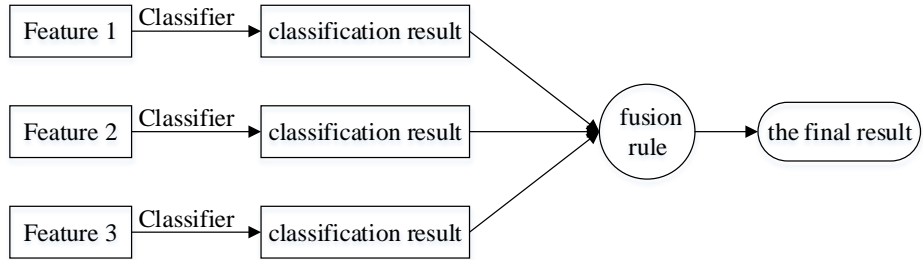

**Fig. 2** Sketch of the high-level features fusion method.

About the classifier, SVM (support vector machine) is a statistical learning algorithm based on the principles of structural risk minimization proposed by Vipnik.[20] The optimal partition of the training sample sets is achieved through the construction of the optimal hyper-plane with maximum interval in the feature space. So we perform image classification by using SVM as a classifier.

According to the above, in this paper we adopted the classification method based on the high-level feature fusion algorithm and SVM, and its core idea is described as follows. First the classification model based on each of the three features (PHOW, PHOC and PHOG) is separately established by using support vector machines.[21] Then the classification results that correspond to each of the three feature are assigned weightings which is adjusted by the process of learning or training. Finally, according to classification model based on each of the three features obtained in the training set, compute the probabilities based on each of the three features that every image in the testing set belongs to each category, which are coupled with the weightings of the three features to compute the final probabilities that every image in the testing set belongs to each category.[22] The detailed steps of the algorithm are illustrated:

(1) Model training: For the training sample set with images of *N* categories, the classifiers based on PHOC, PHOG and PHOW features are respectively trained by SVM expressed as PHOW-classifier, PHOC-classifier and PHOG-classifier;

(2) Feature weight learning: Initializes $a_{PHOW}$, and $a_{PHOG}$ to zero. The trained PHOW-classifier, PHOC-classifier and PHOG-classifier and in the step (1) classifier are separately used in the recognition of every image in the training sample set. If the PHOW-classifier shows the correct category information, then make $a_{PHOW}++$; If the PHOC-classifier shows the correct category information, then make $a_{PHOC}++$; If the PHOG-classifier shows the correct category information, then make $a_{PHOG}++$. With all the images in the training sample set have been recognized, the resulting $a_{PHOW}$, $a_{PHOC}$ and $a_{PHOG}$ and are expressed as the weight of each of the three features after normalization.

(3) Test algorithm: According to the three kinds of features of the test image, PHOW-classifier, PHOC-classifier and PHOG-classifier are used to compute the probability of the test image belonging to each category respectively which is expressed as $D_j(PHOW)$, $D_j(PHOC)$ and $D_j(PHOG)$, $j=1,2,\cdots,N$. And then the classification results based on each of the three features are assigned with corresponding weights obtained from the step (2) in a cumulative way. That is, the probability of the image belonging to each category can be calculated in accordance with the formula:

$$D_j = a_{PHOC} \times D_j(PHOC) + a_{PHOG} \times D_j(PHOG) + a_{PHOW} \times D_j(PHOW), j=1,2,\cdots,N \quad (6)$$

Finally, the category information of the test image can be determined according to the maximum value of $D_j$, that is, $Label = \text{argmax}\{D_1, D_2, \cdots D_N\}$.

## 4 Experiments and Discussions

*4.1 Experimental database and parameter settings*

In the experiment a total of 8 categories of the images were used and each category was composed of 120 images in which 60 images are as the training set and the rest 60 images are as the testing set. The images are mainly from the Caltech101 standard database. Figure 3 shows the experimental images and the sample images.

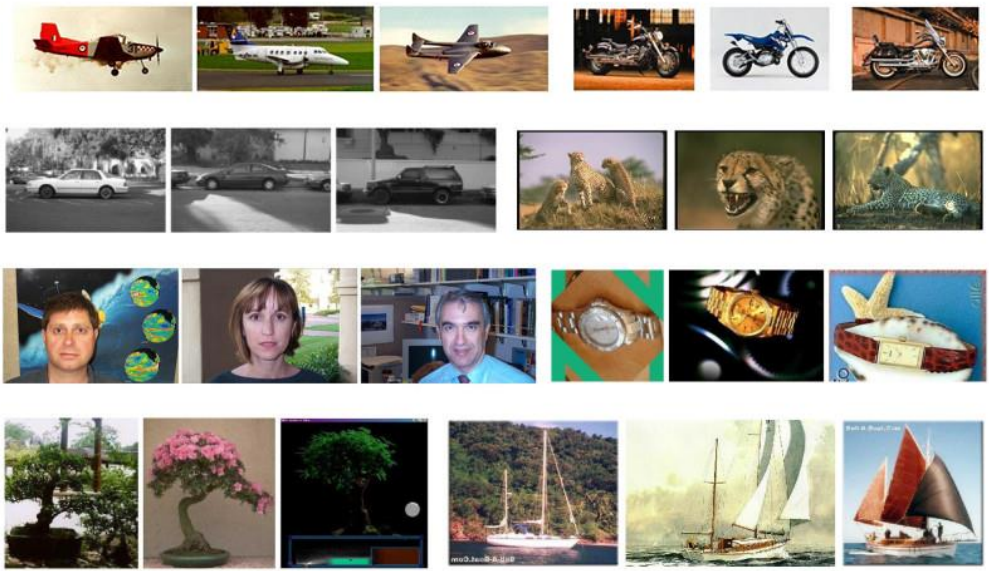

**Fig. 3** Example images from the Caltech-101 dataset.

According to Sec. 2, each image can be described with three features containing PHOW, PHOC and PHOG, and each of the three features has $L+1$ kinds of description by spatial pyramid multi-resolution decomposition (*L* is the highest level of partition), so $3(L+1)$ feature descriptors can be extracted from each image. In this paper the highest level of partition *L* is taken as 2, as a resuit a total of nine feature descriptors can be obtained.

About the PHOW feature extraction, sampling interval is set to eight pixels in the process of dense sampling, each of the 16 × 16 pixel blocks is represented by a feature vector of SIFT descriptor with 128-dimensional. The size of the visual vocabulary is set to 200, eventually the dimension of PHOW feature equals to 200 * (1 + 4 + 16) = 4200. In the process of PHOC feature extraction, for each sub-block of the image pyramid at all levels, the color values are represented by a histogram with 8 bins for H, 3 bins for S, and 3 bins for V, giving a 72 bin histogram for each sub-block. By combining the histograms of all sub-blocks of the image pyramid at all levels, each image can be represented by a PHOC feature descriptor with a total of 72 * (1 + 4 + 16) = 1512 bin. In the process of PHOG feature extraction, the gradient direction is divided into 20 intervals, and the PHOG descriptor can be connected with the three gradient direction histogram vectors (*l* = 0,1,2) together in a sequential order, that is, each image can be represented by a PHOG feature descriptor with the dimension of 20 * (1 + 4 + 16) = 420.

*4.2 Experimental results*

For each category of the images, the three feature representation techniques mentioned above are adopted to train and study the training set respectively in order to get its corresponding SVM classifier differentiating it from other categories. Then the three classifiers are fused into a complex classifier by using the weight coefficients of three features obtained by cross validation.

Experimental study with the classification method as shown in Ref.4 (The image classification based on bag-of-words) and Ref.7 (image classification based on PBOW) as well as this article were conducted to The effectiveness of the algorithm proposed in this paper are confirmed.

The confusion matrix[23, 24] is used to show the classification accuracy about the experimental simulation results, as shown in Fig. 4. The X-axis and Y-axis of the confusion matrix respectively denote the categories of the image. The element in the ith row and the jth column of confusion matrix shows the accuracy of classifying the ith category of the image into the the jth category. The diagonal element of the confusion matrix represents the classification accuracy of each category itself. For the 480 test images, Table 1 shows comparative data of the classification results for the method as shown in Ref. 4 (BOW), Ref. 7 (PBOW) and this article.

By the experimental results it can be seen that the method used in the paper has limited increase of classification accuracy for the images with rich local features such as car_side and leopards while it improves the accuracy greatly for the images with obvious global features as motorbike and watch compared with the method based on PHOW. The classification method based on the fusion of complementary features proposed in this paper has better classified effect than the method based on BOW or PHOW. But overall, in this paper, the classification performance of the PHOW method based on improved BOW is more accurate than that of the traditional PHOW by Ref. 9. The classification performance of the feature fusion algorithm based on the complementary features is superior to that using BOW or PHOW feature. The local information of the image is described by the PHOW feature, while the global information is described by the PHOG and PHOC feature. The three features have good complementarity.

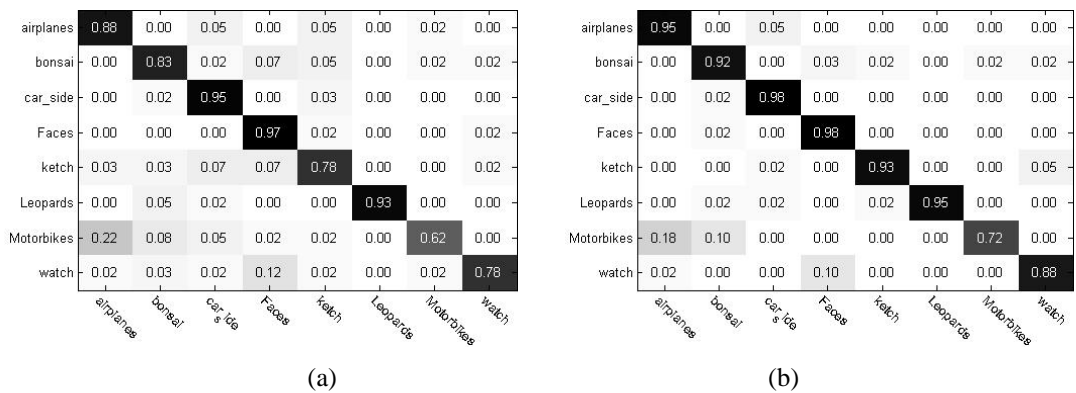

(a)           (b)

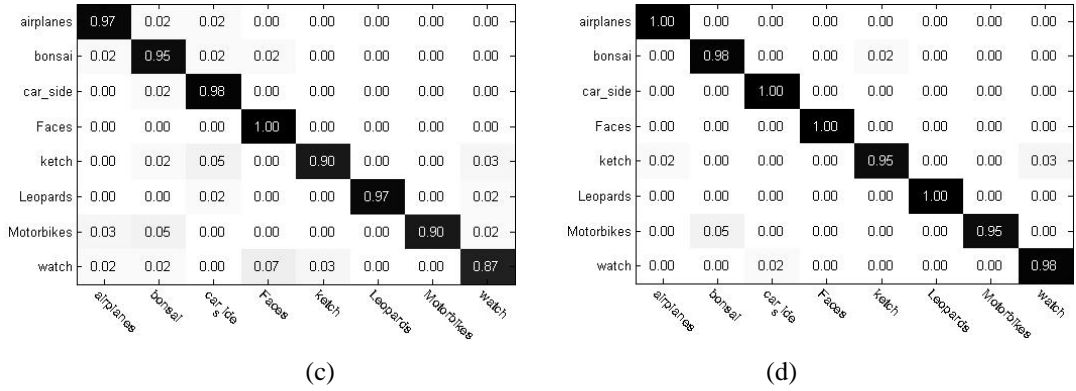

(c)                                      (d)

**Fig. 4** Classification results: (a) Confusion matrix for BOW feature by Ref. 4, (b) Confusion matrix for PHOW Feature by Ref. 9, (c) Confusion matrix for PHOW feature baesd on improved BOW (d) Confusion matrix for PHOW & PHOG & PHOC feature fusion.

**Table 1** Classification performance comparison between the proposed algorithm and the existing algorithms based on the references.

| Algorithm | Correctly classified images | Average accuracy |
|---|---|---|
| Reference[5](BOW) | 405/480 | 84.375% |
| Reference[7](PHOW) | 439/480 | 91.458% |
| PHOW based on improved BOW | 452/480 | 94.167% |
| PHOW&PHOG&PHOC | 472/480 | 98.333% |

In addition, an obvious advantage of the method in this paper is that the weight of each feature can be learned automatically by the training sample set, so that the method has certain adaptive and self-adjustment ability for training and recognition of different categories of images. In the process of feature weight learning, the feature of PHOW model can reach an better classification effect and get a greater weight by cross-validation for the image with abundant local features; The feature of PHOG model can reach an better classification effect and get a greater weight by cross-validation for the image with obvious edge features and shape features; The feature of PHOC model can reach an better classification effect and get a greater weight by cross-validation for the image with obvious global features. Then, the classification result based on one feature can be made up for each other by feature fusion, the probability of error can be reduced, the classification accuracy can be improved.

## 5   Conclusions

This paper proposed an image classification algorithm based on SVM and multi-feature fusion. Integration method of pyramid histogram of color (PHOC), pyramid histogram of oriented gradient (PHOG) and pyramid histogram of words (PHOW) is employed in the process of image feature extraction and description. PHOC and PHOG are based on color feature and gradient direction of the image respectively. The SIFT feature based on dense grid patches combined with improved K-means clustering algorithm is adopted for the image PHOW representation. Both the global feature and the local information of the image were taken into consideration, and it is more complete and more flexible to describe the image feature by the pyramid structure of image space multi-resolution decomposition, so that the performance of image classification can be improved. On the basis of SVM and the above three features fusion method, this article applies a new image classification algorithm based on the adaptive feature-weight adjustment. First the classification model based on each of the three features (PHOC, PHOG and PHOW) is separately established by using support vector machines. Then the weight coefficient that corresponds to each of the three features is adjusted by the process of learning or training according to the training sample set. Finally, the three classifiers are combined into one complex classifier by the weight coefficient of each features, that is, all the decision-making information is fused together to make a common decision. Therefore, the final classification result is obtained by the high-level feature fusion of the three classification results. From the experimental results can be seen that the classification accuracy rate of the proposed method is improved by 7%-17% higher than that of the traditional BOW methods. Through experiments it is proven that this algorithm can effectively combine the global features with the local spatial features of images and significantly improves the accuracy of image classification.

*Acknowledgments*

The presented research work is supported by Foundation for Innovative Research Groups of the National Natural Science Foundation of China (Grant No.61321002), Projects of Major International (Regional) Jiont Research Program NSFC (61120106010), Beijing Education Committee Cooperation Building Foundation Project and Program for Changjiang Scholars and Innovative Research Team in University under Grant IRT1208.

*References*

1. Vailaya, A., Figueiredo, M.A.T. and Jain, A.K., "Image Classification for Content-Based Indexing," *IEEE Transactions on Image Processing*. **10**, 117-130 (2001).
2. Youna J., Eenjun H. and Wonil K. "Sports Image Classification through Bayesian Classifier," *Lecture Notes in Computer Science*, ISSU 3040; pp. 546-555 (2004).

3. Chang E. and Sychay G. "CBSA: Content-Based Soft Annotation for Multimodal Image retrieval using Bayes Point Machines," *IEEE Transactions on Circuits and systems for Video Technology*. **13**, 26-38 (2003).

4. Csurka, G., Bray, C., Dance, C. and Fan, L. "Visual categorization with bags of keypoints. Proceedings of the Workshop on Statistical Learning in Computer Vision," *Proceedings of the 8$^{th}$ European Conference on Computer Vision*, Prague, Czech (2004).

5. Lowe D G. "Object recognition from local scale-invariant features," *Proceedings of the International Conference on Computer Vision*, Corfu, Greece, 1150-1157 (1999).

6. Lowe D G. "Distinctive image features from scale-invariant keypoints," *International Journal of Computer Vision.* **60**(2), 91-110 (2004).

7. Lazebnik S., Schmid C. and Ponce J. "Beyond bags of features: spatial pyramid matching for recognizing natural scene categories," *Proceedings of the IEEE Computer Society Conference of Computer Vision and Pattern Recognition*, New York, USA, June 17-22, 2169-2178 (2006).

8. Xin Z., Bing-quan L. and De-yuan Z. "Study of image classification with spatial pyramid color histogram," *Computer Engineering and Applications*. **46**(18), 152-155(2010).

9. Bosch A., Zisserman A. and Munoz X. "Representing shape with a spatial pyramid kernel," *International Conference on Image and Video Retrieval*, Amsterdam, Netherlands, July 9-11, 401-408 (2007).

10. Grauman K. and Darrell T. "The pyramid match kernel: Discriminative classification with sets of image features," *Proceedings of International Conference on Computer Vision*, 1458-1465 (2005).

11. Bosch A., Muoz X. and Marti R. "Which is the best way to organize/classify images by content?" *Image and Vision Computing*. **25**(6), 778-791 (2006).

12. Jurie F. and Triggs B. "Creating efficient codebooks for visual recognition," *Proceedings of the Tenth IEEE International Conference on Computer Vision*, Beijing, China, 604-610 (2005).

13. Sivic J. and Zisserman A. "Video Google: A text retrieval approach to object matching in videos," *IEEE International Conference on Computer Vision* (2003).

14. Nowak E., Jurie F. and Triggs B. "Sampling strategies for bag-of-features image classification," *Proc of European Conference on Computer Vision*, 490-503 (2006).

15. Chunhui Z., Ying W. and Masahide K. "Improved k-means clustering method for codebook generation," *Chinese Journal of Scientific Instrument*. **33**(10), 2380-2386 (2012).

16. Bosch A., Zisserman A. and Munoz X. "Image classification using random forests and ferns," *IEEE 11th International Conference on Computer Vision*, Rio de Janeiro, Brazil; IEEE Press, 1-8 (2007).

17. Panchanathan S., Park Y.C. and Kim K.S. "The Role of Color in Content-Based Image Retrieval," *International Conference on Image Processing*. 517-520 (2000).

18. Dala N. and Triggs B. "Histograms of oriented gradients for human detection," *IEEE Computer Society Conference on Computer Vision and Pattern Recognition*, San Diego, CA, USA, June 25-25, 886-893 (2005).

19. Jin L., Quan L. and Qingsong A. "Research of image classification based on fusion of SURF and global feature," *Computer Engineering and Applications*. **49**(17), 174-177 (2013).

20. Vapnik V.N. "The Nature of Statistical Learning Theory," New York: Springer (1995).

21. Chang C C. and Lin C J. "Libsvm: a library for support vector machines[EB/OL]," http://www.csie.ntu.edu.tw/~cjlin/libsvm (2001).

22. Yu-feng L., De-quan Z. and Tie-jun Z. "Image categorization based on SVM and the Fusion of multiple features," *The Fourth National Conference of Information Retrieval and Content Security* (2008).

23. Piji L. and Jun M. "What is Happening in a Still Picture?" *2011 First Asian Conference on Pattern Recognition* (2011).

24. Piji L., Jun M. and Shuai Gao. "Actions in Still Web Images: Visualization, Detection and Retrieval," *The 12th International Conference on Web-Age InformationManagement*. Springer (2011).